\documentclass[runningheads]{llncs}
\usepackage{graphicx}
\usepackage{hyperref}
\usepackage{fontawesome}
\usepackage{algpseudocode}
\usepackage{float}

\usepackage{tabularray}
\usepackage{caption}
\usepackage[hmargin=3.7cm]{geometry}

\makeatletter
\newcommand{\github}[1]{%
   \href{#1}{\faGithubSquare}%
}
\makeatother

\hypersetup{
    colorlinks = true,
    citecolor = blue,
    linkcolor = blue,
    urlcolor=blue,
    }
\begin{document}
\title{PoseSync: Robust pose based video synchronization \thanks{Supported by Infocusp Innovations LLP}}

\author{Rishit Javia \and Falak Shah \and
Shivam Dave
}

\institute{Infocusp Innovations LLP \\
\email{\{rishit, falak, shivam\}@infocusp.com}\\
\url{https://infocusp.com/}\\
% \and
% ZZZZ\\
}
\maketitle              % typeset the header of the contribution

\begin{abstract}
Pose based video sychronization can have applications in multiple domains such as gameplay performance evaluation, choreography or guiding athletes. The subject's actions could be compared and evaluated against those performed by professionals side by side. In this paper, we propose an end to end pipeline for synchronizing videos based on pose. The first step crops the region where the person present in the image followed by pose detection on the cropped image. This is followed by application of Dynamic Time Warping(DTW) on angle/ distance measures between the pose keypoints  leading to a scale and shift invariant pose matching pipeline.  

\end{abstract}
\section{Introduction}

Video sychronization task refers to time aligning the frames from multiple videos where
the persons in both videos are trying to perform same action but there are some mismatches in timing and action. This task that is quite intuitive for humans poses a number of challenges as an automated synchronization task, few of which are listed below: 
\begin{itemize}
    \item Pose differences between persons performing the action
    \item Speed difference: would lead to difference in timing of action movements
    \item Scale difference: depending on distance between the person and camera and also inherent size difference
    \item Shift in position of persons within the frame
\end{itemize}

We introduce a tool \textbf{PoseSync} that synchronizes any two videos by bringing them in sync using the state of the art models at its backend for performing pose estimation and matching the poses. It consists of three stages:
\begin{itemize}
    \item Video frame cropping
    \item Pose detection
    \item Video syncronization: using DTW
\end{itemize}

PoseSync, first, crops the video-frames using YOLO v5\cite{yolov5} (we also experimented with tracking using Multiple Instance Learning tracker \cite{miltracker} from OpenCV \cite{tracker} for faster cropping). Cropping operation on the original frames improves the accuracy of pose detection by getting rid of other people in the background/ any spurious information. These cropped frames are passed to pose detection model called MoveNet that returns the pose keypoints for each frame. Finally, Dynamic Time Warping (DTW)\cite{berndt1994using} is used to map the keypoints for both videos (using distance or angle based metrics described later) and map the test video to reference video.  DTW, originally proposed for speech recognition is a general purpose algorithm that can measure the similarity of patterns across different time series. To solve the issue of size differences between two poses, we propose an Angle-Mean Absolute Error metric that computes the MAE between angles of key skeleton joints. This metric is invariant to scale, position and angle of the pose.

Open source implementation of the proposed algorithm can be found \href{https://github.com/InFoCusp/posesync} {here}.

\subsection{Relevant past work}\label{sec:rpw}

Different pose detection models have been proposed in the literarure for detecting keypoints in human poses \cite{MoveNet} \cite{yang2021transpose} \cite{unipose} from an image. TransPose \cite{yang2021transpose} consists of a CNN feature extractor, a Transformer Encoder, and a prediction head. Transformer's attention layers can capture long range spatial relationships in the image that are key to detecting pose keypoints. And the prediction head detects the precise locations of the keypoints by aggraegating heatmaps generated by the transformer. UniPose \cite{unipose} is a single stage pose detection model that utilizes Waterfall Atrous Spatial Pooling (WASP) module proposed by the authors. They obtain a large effective field of view (and multi-scale representaions)  using dilated convolution \cite{dilconv} layers bunched together using “Waterfall” configuration.  Antother human pose detection model, MoveNet is neural network based architecture built to track human pose in real-time from video clips. We will further discuss this model in-depth in section \ref{pose_detection}.

To find the similarity and relationship between two time series, various methods like cross-correlation, dynamic time warping (DTW) have been applied. Utpal Kumar et al. \cite{dtw_seismic} concluded that DTW efficiently captures valueable information which helps to detect even minor variations in time series that windowed cross correlation (WCC) \cite{wcc} fails to catch. 

 In the field of water distribution network, Seubli Lee et al. \cite{dtw_water} found that Dynamic Time Warping (DTW) algorithm performs better in searching for the minimum distance between two water data streams by  comparing different time steps, rather than applying the Euclidean algorithm, which evaluates the data at same time step. Rao et al. \cite{dtw_3d} proposed a DTW aided view-invariant similarity measure to determine temporal correspondence between two videos. Dexter et al. \cite{dexter2009multi} took an alternative approach for video matching, they computed self-similarity matrices to describe the features along the image sequence, and then used these view-invariant descriptors for temporal alignment. \cite{giusti2013empirical} compares 48 dissimilarity metrics empirically to classify various time-series and found that DTW-based metrics outperform the rest. 

Our main contributions are as follows: designing an end to end pose based video synchronization model by putting together the building blocks from different domains. We also introduce a metric for comparing the pose keypoints that is a) invariant to rotation/ translation/ scaling and b) gives more weightage to certain keypoints / joints based on specific task requirements.
 
\section{Pose detection}\label{pose_detection}
Movenet \cite{MoveNet} is a deep learning architecture specifically built for accurately detecting and tracking human poses in real-time from video streams. It is optimized to efficiently operate on mobile devices with constrained computational resources, achieving high frame rates during execution.

It is a bottom-up estimation model which utilizes heatmaps to precisely locate keypoints on the human body. The model comprises of two main components: a feature extractor and a group of prediction heads similar to  CenterNet \cite{centernet}.

It utilizes MobileNetV2 \cite{mobilenetv2} as its feature extractor, which is enhanced with a feature pyramid network (FPN) \cite{fpn}. This combination enables the model to generate semantically rich feature maps with a high resolution output. The feature extractor in MoveNet is accompanied by four prediction heads that are responsible for densely predicting the following:
\begin{itemize}
    \item Person center heatmap: This head predicts the geometric center of individual person instances.
    \item Keypoint regression field: It infers the complete set of keypoints for each person individually, that helps in grouping keypoints into individual instances.
    \item Person keypoint heatmap: This head infers the specific location of all keypoints, regardless of the person instances.
    \item 2D per-keypoint offset field: It predicts local offsets from each pixel in the output feature map to accurately determine the location of each keypoint.
\end{itemize}

\section{Dynamic time warping}\label{dtw}
Dynamic Time Warping (DTW) is a method used to calculate the similarity between two time series. The primary goal of DTW is to identify corresponding matching elements in the time series and measure the distance between them. It relies on dynamic programming principles to determine the optimal temporal alignment between elements in two time series \cite{berndt1994using}. Researchers have successfully applied DTW for analyzing diverse types of sequential data, including audio, video or financial time series. Essentially, any form of data that can be represented as a linear sequence can be effectively analyzed using DTW.

DTW assumed that the below conditions stand true for both sequences:

\begin{itemize}
    \item The first index from the first sequence must be matched with the first index from the other sequence (although it may have additional matches).
    \item The last index from the first sequence must be matched with the last index from the other sequence (while allowing for other matches).
    \item Each index from the first sequence must be matched with one or more indices from the other sequence, and vice versa. 
    \item The mapping of indices from the first sequence to indices from the other sequence must be strictly increasing. This means that if index j comes after index i in the first sequence, there should not be two indices l and k in the other sequence such that index i is matched with index l and index j is matched with index k.
\end{itemize}

We use DTW to syncronize the sequences of human pose detected from two videos. The elements of the sequences are set of keypoints which are used to compute the cost between any two elements from each series. The cost can be computed using metrics like simple mean absolute error or mean absolute error between the angles derived from the keypoints. The metric computation is covered in depth in section \ref{Pose_matching_metric}. 

\section{Pose matching metric}\label{Pose_matching_metric}

Pose matching can be performed between two unique poses using pose keypoints (human body joint : x,y coordinates) as vector representations and using Mean Absolute Error(MAE) or Mean Squared Error(MSE) as distance metrics. The limitation with these metrics is that they are not scale, rotation and shift invariant. That is: even when poses are similar, MAE or MSE between pose keypoints could be high due to scale or position difference. To overcome this problem, we use angle based mean absolute error. It works by first calculating the joint angles formed by three joint points with one joint as pivot.  We then calculate MAE between the angles as distance metric.\\

We use below mentioned 9 joints triplets for angle calculation. We found that these joints are sufficient for most common activities like dancing, exercise, etc:

\begin{itemize}
    \item \textbf{left shoulder joint} : left shoulder, left shoulder,  left elbow
    \item \textbf{right shoulder joint} : right hip, right shoulder, right elbow
    \item \textbf{right elbow joint} : right shoulder, right elbow, right wrist
    \item \textbf{left elbow joint} : left shoulder,left elbow, left wrist
    \item \textbf{right hip joint} : left hip, right hip, right knee
    \item \textbf{left hip joint} : right hip, left hip, left knee
    \item \textbf{right knee joint} : right hip, right knee, right ankle
    \item \textbf{left knee joint} : left hip, left knee, left ankle
    \item \textbf{waist joint} : left shoulder, left hip, left knee   
\end{itemize}

% We further investigate that the angle range of 180 degree in cosine similarity can impose limitation in mathching accuracy. The reason behind this is that the angle computation using cosine can capture angle in the span of semi circle which covers half of the direction in two dimensions. This issue can solved by expanding the angle range to 360 degree. We expand this range by mapping the 2d vector into 3d space by adding z component with coefficient zero into vectos and utilize determinant calculation of this two vectors with principal component vector(unit vector in z direction). The sign of the determinant determines the position of plane formed by two pair of vectors in either part of circle. A negative determinant indicates presence of plane of vectors in the range of 180 to 360 degree. Where as positive sign indicates presence of plane of vectors in the range of 0 to 180 degree. This way resultant angle is 360 - cosine angle if the determinant is less than zero otherwise just cosine angle.\\

We use angle MAE as the cost function for Dynamic Time Warping. Depending on the application, we can also give weightage to different joints/ keypoints and that can be helpful in better synchronization of the given videos.

\section{Results} \label{sec:results}

\begin{figure}
    \centering
    \includegraphics[width=0.8\textwidth,height=0.7\textheight,keepaspectratio]{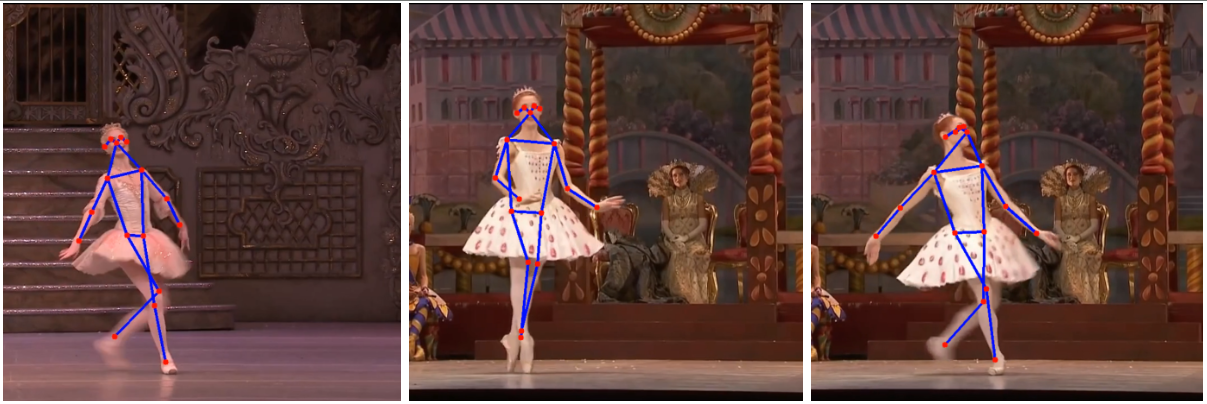}
    \includegraphics[width=0.8\textwidth,height=0.7\textheight,keepaspectratio]{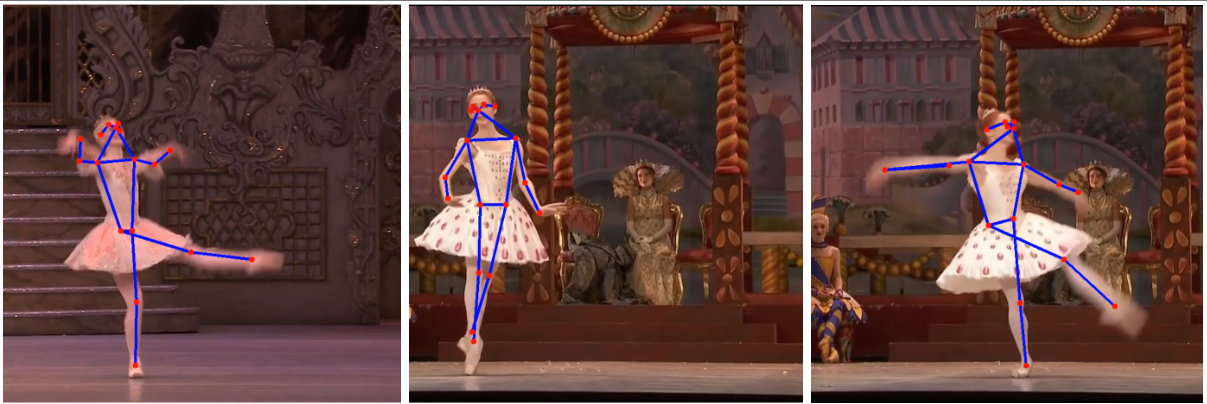}
    \includegraphics[width=0.8\textwidth,height=0.7\textheight,keepaspectratio]{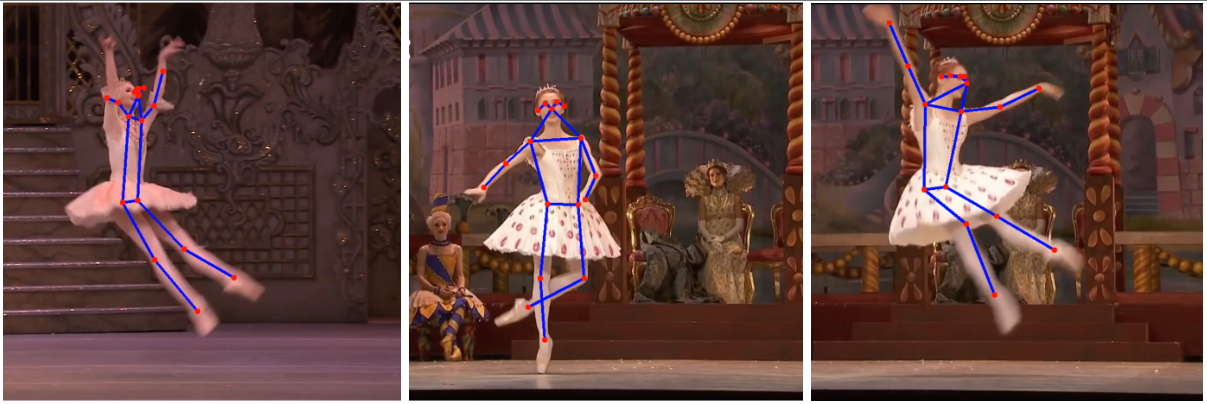}
    \includegraphics[width=0.8\textwidth,height=0.7\textheight,keepaspectratio]{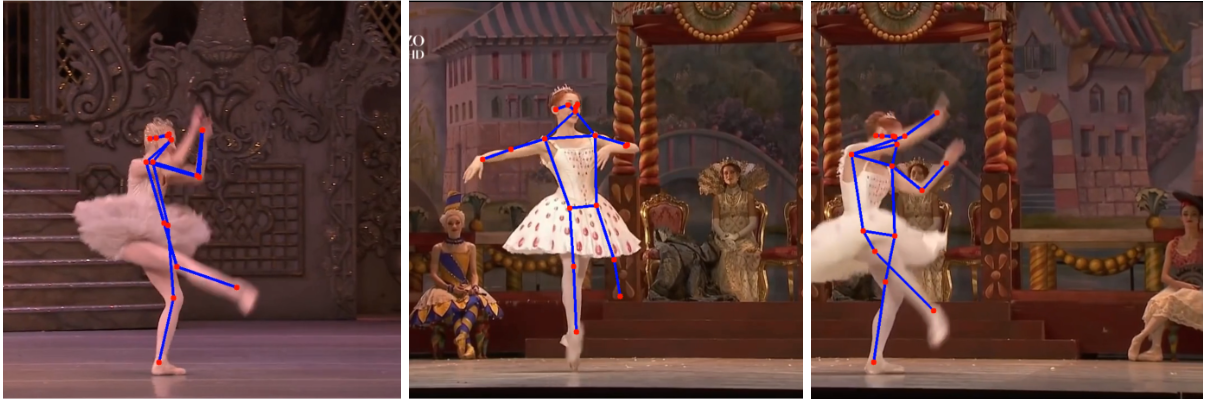}
    \includegraphics[width=0.8\textwidth,height=0.7\textheight,keepaspectratio]{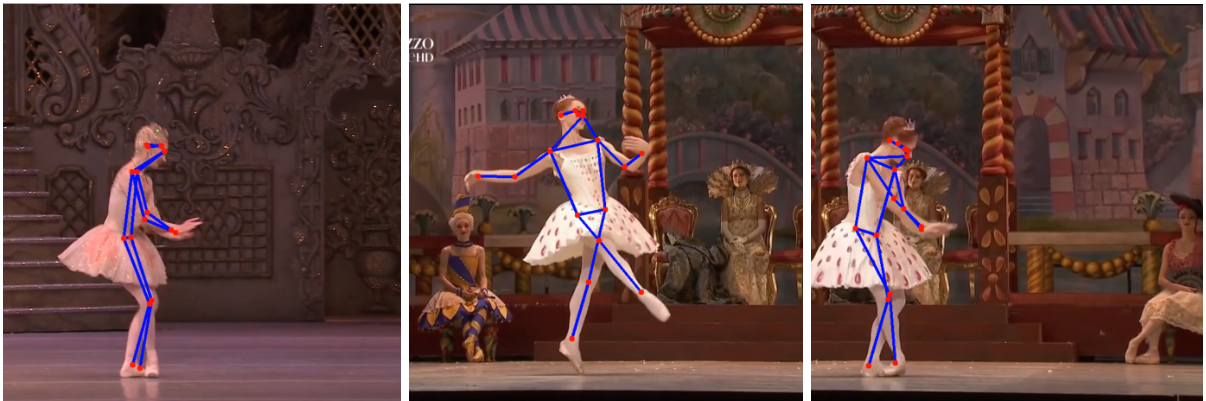}
    \caption{Dance videos synchronization : Key frames of reference video (column 1), corresponding test video frames (column 2) and test video frames mapped to respective reference frames by DTW (column 3)}
    \label{fig:dance}
\end{figure}

\begin{figure}
    \centering
    \includegraphics[width=0.8\textwidth,height=0.7\textheight,keepaspectratio]{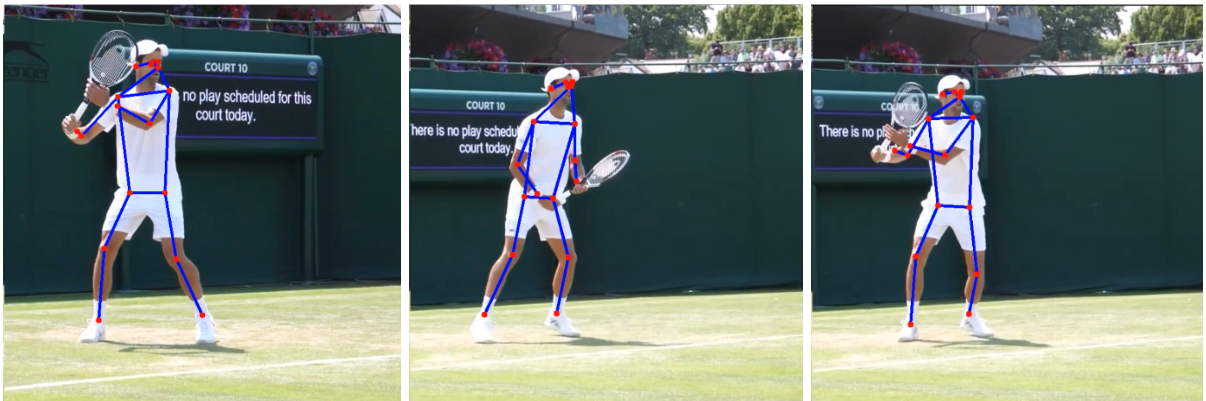}
    \includegraphics[width=0.8\textwidth,height=0.7\textheight,keepaspectratio]{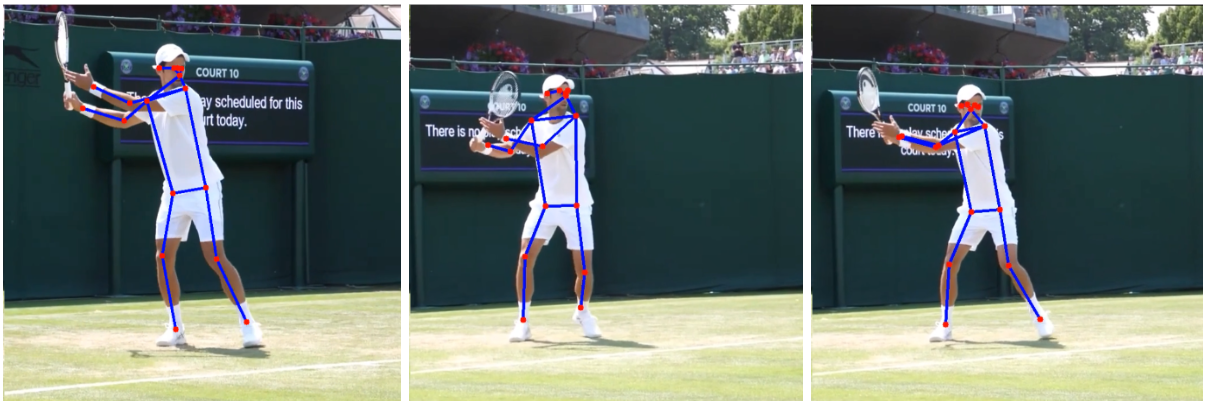}
    \includegraphics[width=0.8\textwidth,height=0.7\textheight,keepaspectratio]{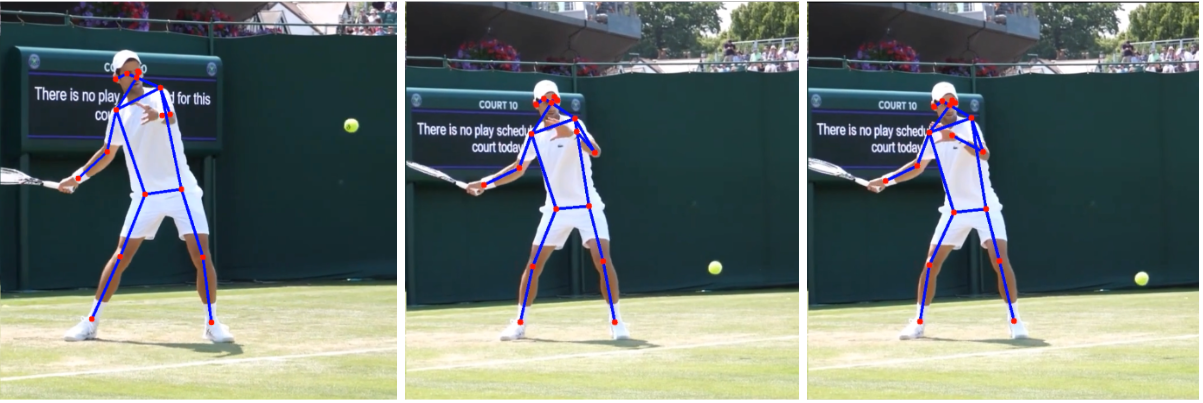}
    \includegraphics[width=0.8\textwidth,height=0.7\textheight,keepaspectratio]{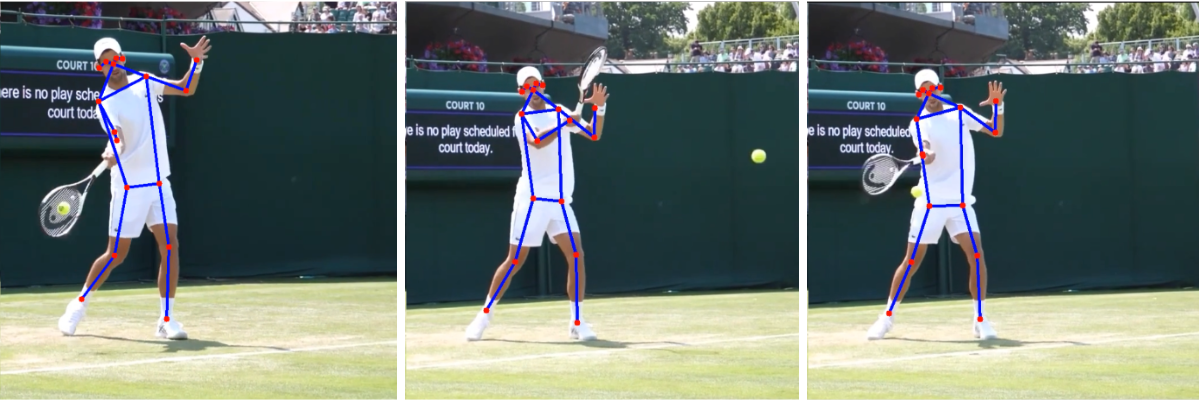}
    \includegraphics[width=0.8\textwidth,height=0.7\textheight,keepaspectratio]{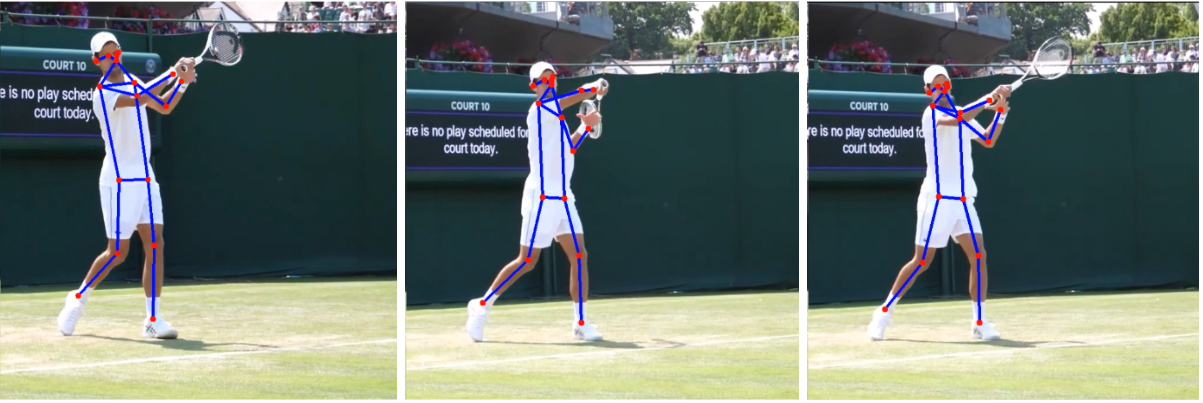}
    \caption{Tennis shots synchronization : Key frames of reference video (column 1), corresponding test video frames (column 2) and test video frames mapped to respective reference frames by DTW (column 3)}    
    \label{fig:tennis}
\end{figure}

We applied our algorithm, PoseSync on various videos for temporal alignment between actions. The alignment of videos, containing human activities illustrates the robustness of the DTW aided algorithm. Figure \ref{fig:dance} and \ref{fig:tennis} show the video alignment between two videos with different type of human movements. First column consists of reference video key frames and second column contains test video frames at the same index as the reference frames. The third column consists of test frames mapped to reference frames using PoseSync. Results show that it can map the similar poses accurately in videos with similar activities.

 We used various video combinations such as: original video and video with some noise (different clip of same action) or clip of different action or increased/decreased speed. Test videos are generated by increasing/decreasing speed of entire reference video or beginning/middle/end part, or putting other video clip of same action or different action in the start/middle/end. PoseSync can match the video very well even if other video has noise of 2 sec anywhere in the clip of 10 seconds. In case of different speed of both videos, it is able to syncronize the videos with good accuracy as shown in Table \ref{table:table1}.
 \\

% \begin{figure}{H}
%     \centering
%     \includegraphics[width=0.8\textwidth,height=0.7\textheight,keepaspectratio]{w_djo.png}
%     \includegraphics[width=0.8\textwidth,height=0.7\textheight,keepaspectratio]{w_fed_ref_eq_dyn.png}
    
%     \caption{Adaptive Metrics : Column 1 shows the reference video frame. Column 2 and 3 represent  mapped test frames with equal and dynamic keypoints weightage, respectively.}
%     \label{fig:metrics}
% \end{figure}

% Improved pipeline metrics:
\begin{table}[H]
\begin{tabular}{|cc|cc|ccc|}
\hline
\multicolumn{2}{|c|}{Reference Video}                                                                                                                                                 & \multicolumn{2}{c|}{Test Video}                                                                                                                                                                                      & \multicolumn{3}{c|}{Video Matching}                                                                                                                                                                                                                                 \\ \hline
\multicolumn{1}{|c|}{\begin{tabular}[c]{@{}c@{}}Length \\ (in sec)\end{tabular}} & Description                                                                                        & \multicolumn{1}{c|}{\begin{tabular}[c]{@{}c@{}}Length \\ (in sec)\end{tabular}} & Description                                                                                                                        & \multicolumn{1}{c|}{\begin{tabular}[c]{@{}c@{}}No. of frames \\ expected \\ to match\end{tabular}} & \multicolumn{1}{c|}{\begin{tabular}[c]{@{}c@{}}No. of frame \\ matched \\ actually\end{tabular}} & \begin{tabular}[c]{@{}c@{}}\% Video \\ matched\end{tabular} \\ \hline
\multicolumn{1}{|c|}{1}                                                          & Sample video                                                                                       & \multicolumn{1}{c|}{1}                                                          & Same sample video                                                                                                                  & \multicolumn{1}{c|}{25}                                                                            & \multicolumn{1}{c|}{25}                                                                          & 100                                                         \\ \hline
\multicolumn{1}{|c|}{7}                                                          & \begin{tabular}[c]{@{}c@{}}action clips(A,B), \\ each of 2 sec, \\ ordered as A\_B\_A\end{tabular} & \multicolumn{1}{c|}{7}                                                          & \begin{tabular}[c]{@{}c@{}}action clips(A,B), \\ each of 2 sec, \\ ordered as B\_A\_B\end{tabular}                                 & \multicolumn{1}{c|}{163}                                                                           & \multicolumn{1}{c|}{102}                                                                         & 62.57668712                                                 \\ \hline
\multicolumn{1}{|c|}{12}                                                         & \begin{tabular}[c]{@{}c@{}}action clips(A,B), \\ each of 6 sec, \\ ordered as A\_B\end{tabular}    & \multicolumn{1}{c|}{12}                                                         & \begin{tabular}[c]{@{}c@{}}action clips(A,B), \\ each of 6 sec, \\ ordered as B\_A\end{tabular}                                    & \multicolumn{1}{c|}{150}                                                                           & \multicolumn{1}{c|}{130}                                                                         & 86.66666667                                                 \\ \hline
\multicolumn{1}{|c|}{8}                                                          & Normal video                                                                                       & \multicolumn{1}{c|}{10}                                                         & \begin{tabular}[c]{@{}c@{}}Normal video(0 - 4) sec + \\ 2 sec noise + \\ normal video(4 - 8) sec\end{tabular}                      & \multicolumn{1}{c|}{194}                                                                           & \multicolumn{1}{c|}{173}                                                                         & 89.17525773                                                 \\ \hline
\multicolumn{1}{|c|}{8}                                                          & Normal video                                                                                       & \multicolumn{1}{c|}{10}                                                         & \begin{tabular}[c]{@{}c@{}}2 sec noise + \\ normal video (0 - 8) sec\end{tabular}                                                  & \multicolumn{1}{c|}{194}                                                                           & \multicolumn{1}{c|}{179}                                                                         & 92.26804124                                                 \\ \hline
\multicolumn{1}{|c|}{8}                                                          & Normal video                                                                                       & \multicolumn{1}{c|}{10}                                                         & \begin{tabular}[c]{@{}c@{}}normal video (0 - 8) sec + \\ 2 sec noise\end{tabular}                                                  & \multicolumn{1}{c|}{194}                                                                           & \multicolumn{1}{c|}{189}                                                                         & 97.42268041                                                 \\ \hline
\multicolumn{1}{|c|}{8}                                                          & Normal video                                                                                       & \multicolumn{1}{c|}{9}                                                          & \begin{tabular}[c]{@{}c@{}}Normal video (0 - 4) sec + \\ 1 sec noise + \\ normal video (4 - 8) sec\end{tabular}                    & \multicolumn{1}{c|}{194}                                                                           & \multicolumn{1}{c|}{185}                                                                         & 95.36082474                                                 \\ \hline
\multicolumn{1}{|c|}{8}                                                          & Normal video                                                                                       & \multicolumn{1}{c|}{9}                                                          & \begin{tabular}[c]{@{}c@{}}1 sec noise + \\ normal video (0 - 8) sec\end{tabular}                                                  & \multicolumn{1}{c|}{194}                                                                           & \multicolumn{1}{c|}{172}                                                                         & 88.65979381                                                 \\ \hline
\multicolumn{1}{|c|}{8}                                                          & Normal video                                                                                       & \multicolumn{1}{c|}{9}                                                          & \begin{tabular}[c]{@{}c@{}}normal video (0 - 8) sec + \\ 1 sec noise\end{tabular}                                                  & \multicolumn{1}{c|}{194}                                                                           & \multicolumn{1}{c|}{191}                                                                         & 98.45360825                                                 \\ \hline
\multicolumn{1}{|c|}{8}                                                          & Normal video                                                                                       & \multicolumn{1}{c|}{10}                                                         & \begin{tabular}[c]{@{}c@{}}same as normal video + \\ 2 sec noise\end{tabular}                                                      & \multicolumn{1}{c|}{237}                                                                           & \multicolumn{1}{c|}{228}                                                                         & 96.20253165                                                 \\ \hline
\multicolumn{1}{|c|}{8}                                                          & Normal video                                                                                       & \multicolumn{1}{c|}{10}                                                         & \begin{tabular}[c]{@{}c@{}}2 sec noise + \\ same as normal video\end{tabular}                                                      & \multicolumn{1}{c|}{239}                                                                           & \multicolumn{1}{c|}{238}                                                                         & 99.58158996                                                 \\ \hline
\multicolumn{1}{|c|}{8}                                                          & Normal video                                                                                       & \multicolumn{1}{c|}{10}                                                         & \begin{tabular}[c]{@{}c@{}}Normal video (0 - 4) sec + \\ 2 sec noise + \\ normal video (4 - 8) sec\end{tabular}                    & \multicolumn{1}{c|}{239}                                                                           & \multicolumn{1}{c|}{219}                                                                         & 91.63179916                                                 \\ \hline
\multicolumn{1}{|c|}{8}                                                          & Normal video                                                                                       & \multicolumn{1}{c|}{10}                                                         & \begin{tabular}[c]{@{}c@{}}Normal video (0 - 4) sec + \\ 2 sec clip of different action \\ + normal video (4 - 8) sec\end{tabular} & \multicolumn{1}{c|}{239}                                                                           & \multicolumn{1}{c|}{230}                                                                         & 96.23430962                                                 \\ \hline
\multicolumn{1}{|c|}{1}                                                          & Normal video                                                                                       & \multicolumn{1}{c|}{1}                                                          & flipped normal video                                                                                                               & \multicolumn{1}{c|}{25}                                                                            & \multicolumn{1}{c|}{25}                                                                          & 100                                                         \\ \hline
\multicolumn{1}{|c|}{7}                                                          & Normal video                                                                                       & \multicolumn{1}{c|}{9}                                                          & \begin{tabular}[c]{@{}c@{}}clip of $\sim$2 sec slowed \\ down in middle\end{tabular}                                               & \multicolumn{1}{c|}{163}                                                                           & \multicolumn{1}{c|}{160}                                                                         & 98.1595092                                                  \\ \hline
\multicolumn{1}{|c|}{4}                                                          & Normal video                                                                                       & \multicolumn{1}{c|}{9}                                                          & video slowed down                                                                                                                  & \multicolumn{1}{c|}{105}                                                                           & \multicolumn{1}{c|}{104}                                                                         & 99.04761905                                                 \\ \hline
\multicolumn{1}{|c|}{7}                                                          & Normal video                                                                                       & \multicolumn{1}{c|}{10}                                                         & \begin{tabular}[c]{@{}c@{}}clip of $\sim$2 sec slowed \\ down in middle\end{tabular}                                               & \multicolumn{1}{c|}{237}                                                                           & \multicolumn{1}{c|}{235}                                                                         & 99.15611814                                                 \\ \hline
\multicolumn{1}{|c|}{8}                                                          & Normal video                                                                                       & \multicolumn{1}{c|}{10}                                                         & \begin{tabular}[c]{@{}c@{}}clip of $\sim$2 sec slowed \\ down in the end\end{tabular}                                              & \multicolumn{1}{c|}{211}                                                                           & \multicolumn{1}{c|}{210}                                                                         & 99.52606635                                                 \\ \hline
\multicolumn{1}{|c|}{8}                                                          & Normal video                                                                                       & \multicolumn{1}{c|}{9}                                                          & \begin{tabular}[c]{@{}c@{}}clip of $\sim$2 sec slowed \\ down in the start\end{tabular}                                            & \multicolumn{1}{c|}{207}                                                                           & \multicolumn{1}{c|}{199}                                                                         & 96.1352657                                                  \\ \hline
\multicolumn{1}{|c|}{3}                                                          & Normal video                                                                                       & \multicolumn{1}{c|}{7}                                                          & video slowed down                                                                                                                  & \multicolumn{1}{c|}{102}                                                                           & \multicolumn{1}{c|}{102}                                                                         & 100                                                         \\ \hline
\multicolumn{1}{|c|}{3}                                                          & Normal video                                                                                       & \multicolumn{1}{c|}{2}                                                          & video sped up                                                                                                                      & \multicolumn{1}{c|}{102}                                                                           & \multicolumn{1}{c|}{100}                                                                         & 98.03921569                                                 \\ \hline
\multicolumn{1}{|c|}{3}                                                          & Normal video                                                                                       & \multicolumn{1}{c|}{13}                                                         & \begin{tabular}[c]{@{}c@{}}video speed \\ decreased to 25\%\end{tabular}                                                           & \multicolumn{1}{c|}{102}                                                                           & \multicolumn{1}{c|}{102}                                                                         & 100                                                         \\ \hline
\multicolumn{1}{|c|}{7}                                                          & Normal video                                                                                       & \multicolumn{1}{c|}{7}                                                         & \begin{tabular}[c]{@{}c@{}}zoomed in \\ video\end{tabular}                                                           & \multicolumn{1}{c|}{105}                                                                           & \multicolumn{1}{c|}{102}                                                                         & 96.19                                                         \\ \hline
\end{tabular}
\captionof{table}{Accuracy metrics across various scenarios}\label{table:table1}
\end{table}
\section{Conclusion}\label{conclusion}

We propose a method for sychronizing videos using a rotation/ translation/ scaling invariant metric of pose comparison, called PoseSync.
Since MoveNet is limited to detect pose of single person in the image, the video is needed to be cropped first. So video is processed through YOLO v5 or OpenCV Tracker to get the cropped video frames which are passed to pose detection model, MoveNet. This MoveNet model returns 17 keypoints for each frame. Now we have 2 sequences of 17 keypoints as PoseSync takes two videos as input. Then, to sychronize two videos, we pass keypoints sequences to Dynamic Time Warping (DTW) which computes distance based on MAE or/and Angle-MAE between two videos and maps test video to reference video. To solve the issue of size differences between two poses, we use the Angle based metrics, Angle-Mean Absolute Error which computes the MAE between angles of joints and is invariant to scale, position and angle of the pose.

\bibliographystyle{splncs04}
\bibliography{bibliography}

\begin{thebibliography}{10}
\providecommand{\url}[1]{\texttt{#1}}
\providecommand{\urlprefix}{URL }
\providecommand{\doi}[1]{https://doi.org/#1}

\bibitem{unipose}
Artacho, B., Savakis, A.: Unipose: Unified human pose estimation in single
  images and videos. In: Proceedings of the IEEE/CVF conference on computer
  vision and pattern recognition. pp. 7035--7044 (2020)

\bibitem{miltracker}
Babenko, B., Yang, M.H., Belongie, S.: Visual tracking with online multiple
  instance learning. In: 2009 IEEE Conference on computer vision and Pattern
  Recognition. pp. 983--990. IEEE (2009)

\bibitem{berndt1994using}
Berndt, D.J., Clifford, J.: Using dynamic time warping to find patterns in time
  series. In: KDD workshop. vol.~10, pp. 359--370. Seattle, WA, USA: (1994)

\bibitem{wcc}
Boker, S.M., Rotondo, J.L., Xu, M., King, K.: Windowed cross-correlation and
  peak picking for the analysis of variability in the association between
  behavioral time series. Psychological methods  \textbf{7}(3), ~338 (2002)

\bibitem{dilconv}
Chen, L.C., Papandreou, G., Kokkinos, I., Murphy, K., Yuille, A.L.: Deeplab:
  Semantic image segmentation with deep convolutional nets, atrous convolution,
  and fully connected crfs. IEEE transactions on pattern analysis and machine
  intelligence  \textbf{40}(4),  834--848 (2017)

\bibitem{MoveNet}
Chen, Y.H., Oerlemans, A., Belletti, F., Bunner, A., Sundaram, V.: {MoveNet}:
  Next-generation pose detection model (2021),
  \url{https://blog.tensorflow.org/2021/05/next-generation-pose-detection-with-movenet-and-tensorflowjs.html}

\bibitem{dexter2009multi}
Dexter, E., P{\'e}rez, P., Laptev, I.: Multi-view synchronization of human
  actions and dynamic scenes. In: BMVC. pp. 1--11 (2009)

\bibitem{centernet}
Duan, K., Bai, S., Xie, L., Qi, H., Huang, Q., Tian, Q.: Centernet: Keypoint
  triplets for object detection. In: Proceedings of the IEEE/CVF International
  Conference on Computer Vision (ICCV) (October 2019)

\bibitem{giusti2013empirical}
Giusti, R., Batista, G.E.: An empirical comparison of dissimilarity measures
  for time series classification. In: 2013 Brazilian Conference on Intelligent
  Systems. pp. 82--88. IEEE (2013)

\bibitem{yolov5}
Jocher, G.: Yolov5 by ultralytics (2020). \doi{10.5281/zenodo.3908559},
  \url{https://github.com/ultralytics/yolov5}

\bibitem{dtw_seismic}
Kumar, U., Legendre, C.P., Zhao, L., Chao, B.F.: {Dynamic Time Warping as an
  Alternative to Windowed Cross Correlation in Seismological Applications}.
  Seismological Research Letters  \textbf{93}(3),  1909--1921 (03 2022),
  \url{https://doi.org/10.1785/0220210288}

\bibitem{dtw_water}
Lee, S., Kim, J., Hwang, J., Lee, E., Lee, K.J., Oh, J., Park, J., Heo, T.Y.:
  Clustering of time series water quality data using dynamic time warping: A
  case study from the bukhan river water quality monitoring network. Water
  \textbf{12}(9) (2020), \url{https://www.mdpi.com/2073-4441/12/9/2411}

\bibitem{fpn}
Lin, T.Y., Dollar, P., Girshick, R., He, K., Hariharan, B., Belongie, S.:
  Feature pyramid networks for object detection. In: Proceedings of the IEEE
  Conference on Computer Vision and Pattern Recognition (CVPR) (July 2017)

\bibitem{tracker}
OpenCV: {TrackerMIL},
  \url{https://docs.opencv.org/3.4/d0/d26/classcv_1_1TrackerMIL.html}

\bibitem{dtw_3d}
Rao, Gritai, Shah, Syeda-Mahmood: View-invariant alignment and matching of
  video sequences. In: Proceedings Ninth IEEE International Conference on
  Computer Vision. pp. 939--945 vol.2 (2003). \doi{10.1109/ICCV.2003.1238449}

\bibitem{mobilenetv2}
Sandler, M., Howard, A., Zhu, M., Zhmoginov, A., Chen, L.C.: Mobilenetv2:
  Inverted residuals and linear bottlenecks. In: Proceedings of the IEEE
  Conference on Computer Vision and Pattern Recognition (CVPR) (June 2018)

\bibitem{yang2021transpose}
Yang, S., Quan, Z., Nie, M., Yang, W.: Transpose: Keypoint localization via
  transformer. In: Proceedings of the IEEE/CVF International Conference on
  Computer Vision. pp. 11802--11812 (2021)

\end{thebibliography}

\end{document}